\documentclass[11pt]{article}
\usepackage{fullpage}
\usepackage{bm,amsmath,amssymb}
\usepackage{custom_style}
\usepackage{graphicx}
\usepackage[linesnumbered, ruled]{algorithm2e}
\usepackage{hyperref}
\usepackage{color}
\usepackage{url}
\usepackage{array}
\definecolor{mydarkblue}{rgb}{0,0.08,0.45}
\hypersetup{ 
    colorlinks,
    citecolor=mydarkblue,
    filecolor=mydarkblue,
    linkcolor=mydarkblue,
    urlcolor=mydarkblue 
}

\newcommand{\mname}{\texttt{Med2Vec}\xspace}
\newcommand{\argmax}{\operatornamewithlimits{argmax}}
\newcommand{\argmin}{\operatornamewithlimits{argmin}}

\begin{document}

\title{\textbf{Multi-layer Representation Learning\\ for Medical Concepts}%
\date{}
\author{
Edward Choi$^*$ \quad Mohammad Taha Bahadori$^*$ \quad Elizabeth Searles$^\dagger$ \quad Catherine Coffey$^\dagger$\\ Jimeng Sun$^*$\\
$^*$ Georgia Institute of Technology\\
$^\dagger$ Children Healthcare of Atlanta
} %
}

\maketitle
\begin{abstract}
Learning efficient representations for concepts has been proven to be an important basis for many applications such as machine translation or document classification. 
Proper representations of medical concepts such as diagnosis, medication, procedure codes and visits will have broad applications in healthcare analytics.  
However, in Electronic Health Records (EHR) the visit sequences of patients include multiple concepts (diagnosis, procedure, and medication codes) per visit. 
This structure provides two types of relational information, namely sequential order of visits and co-occurrence of the codes within each visit. 
In this work, we propose \texttt{Med2Vec}, which not only learns distributed representations for both medical codes and visits from a large EHR dataset with over 3 million visits, but also allows us to interpret the learned representations confirmed positively by clinical experts.
In the experiments, \texttt{Med2Vec} displays significant improvement in key medical applications compared to popular baselines such as Skip-gram, GloVe and stacked autoencoder, while providing clinically meaningful interpretation.
\end{abstract}

%

\section{Introduction}
\label{sec:intro}
Discovering efficient representations of discrete high dimensional concepts has been a key challenge in a variety of applications recently \cite{bengio2013representation}. Using various types of neural networks, high-dimensional raw data can be transformed to continuous real-valued concept vectors that efficiently capture their latent relationship from data. Such succinct representations have been shown to improve the performance of various complex tasks across domains spanning from image processing \cite{lecun1998gradient,hinton2006fast,vincent2008extracting}, language modeling \cite{bengio2003neural,mikolov2010recurrent}, word embedding \cite{mikolov2013distributed,pennington2014glove}, music information retrieval \cite{schluter2014improved}, sentiment analysis \cite{socher2013recursive}, and more recently multi-modal learning of images and text \cite{kiros2014multiplicative}.

Efficient representations for concepts is an important, if not essential, element in healthcare as well. 
Healthcare concepts contain rich latent relationships that cannot be represented by simple one-hot coding \cite[Chapter 2.3.2]{murphy2012machine}. For example, pneumonia and bronchitis are clearly more related than pneumonia and obesity. 
In one-hot coding, such relationship between different codes are not represented. Despite its limitation, many healthcare applications \cite{robert2015athsma, sun_predicting_2014} still use the simple sum over one-hot vectors to derive patient feature vectors. 
To overcome this limitation, it is common in healthcare applications, to rely on carefully designed feature representations~\cite{sun2012supervised,ghassemi2014unfolding,wang2015early}. 
However, this process often involves supervision information and ad-hoc feature engineering that requires considerable expert medical knowledge and is not scalable in general. 

Recently, studies have shown that it is possible to learn efficient representations of healthcare concepts without medical expertise and still significantly improve the performance of various healthcare applications. 
Choi et al. \cite{choi2016medical} learned distributed representations of medical codes (\textit{e.g.} diagnosis, medication, procedure codes) using Skip-gram \cite{mikolov2013distributed} and applied them to heart failure prediction. Choi et al. \cite{choi2016learning} also learned the representations for medical concepts from a medical claims dataset and compared the learned representations to existing medical ontologies and code groupers. 
Despite these progress, learning efficient representations of healthcare concepts, however, is still an open challenge. The difficulty stems from several aspects: 
\begin{enumerate}
\item Healthcare data have a unique structure where the visits are temporally ordered but the medical codes within a visit form an unordered set. A sequence of visits possesses sequential relationship among them which cannot be captured by simply aggregating code-level representations.  Moreover, given the demographic information for patients, the structure becomes more complex. 
\item Learned representations should be interpretable. While the interpretability of the model in the clinical domain is considered to be an essential requirement, some of the state-of-the art representation learning methods such as recurrent neural networks (RNN) are difficult to interpret. 
\item The algorithm should be scalable enough to handle real-world healthcare datasets with millions of patients and hundred millions of visits. 
\end{enumerate}
To address such challenges in healthcare concept representation learning, we propose \mname and make the following contributions.
\begin{itemize}
\item
We propose \mname, a simple and robust algorithm to efficiently learn succinct code-, and visit-level representations by using real-world electronic health record (EHR) datasets, without depending on expert medical knowledge. 
\item
\mname learns interpretable representations and enables clinical applications to offer more than just improved performances. We  conducted detailed user study with clinical experts to validate the interpretability of the resulting representation. 
\item
We conduct experiments to demonstrate the scalability of \mname, and show that our model can be readily applied to near 30K medical codes over two large datasets with  3 million and 5.5 million visits, respectively.
\item
We apply the learned representations to various real-world health problems and demonstrate the improved performance enabled by \mname compared to popular baselines.
\end{itemize}

In the following section, we discuss related works, then describe our method in section 3. In section 4, we explain experiment design and interpretation method in detail. We present the results and discussion in section 5. Then we conclude this paper with future work in section 6.

\section{Preliminaries and Related Work}
\label{sec:related}
In this section, we first describe the preliminary ideas used in learning representation for words. 
Then, we review the algorithms developed for representing healthcare data.

\subsection{Learning representation for words}
Representation learning of words using neural network based methods have been studied since the early 2000's \cite{bengio2003neural,collobert2008unified,mnih2009scalable,turian2010word}.
Among these techniques, Skip-gram \cite{mikolov2013distributed} is the basis of many concept representation learning methods, including our own. Skip-gram is able to capture the subtle relationships between words, thus outperforming the previous works in a word analogy task\cite{mikolov2013efficient}.

Given a sequence of words $w_1, w_2, \ldots, w_T$, Skip-gram learns the word representations based on the co-occurrence information of words inside a context window of a predefined size. 
The key principle of Skip-gram is that a word's representation should be able to predict the neighboring words. 
The objective of Skip-gram is to maximize the following average log probability.
\begin{equation*}
\frac{1}{T} \sum_{t=1}^{T} \sum_{-c \leq j \leq c, j \ne 0} \log p(w_{t+j}|w_t) 
\end{equation*}
where $c$ is the size of the context window. The conditional probability is defined by the softmax function:
\begin{equation*}
p(w_O|w_I) = \frac{\exp \Big({v}_{w_O}'^{\top} v_{w_I} \Big)}{\sum_{w=1}^{W}\exp \Big({v}_{w}'^{\top} v_{w_I} \Big)}
\end{equation*}
where $v_w$ and ${v}_{w}'$ are the \textit{input} and \textit{output} vector representations of word $w$. $W$ is the number of words in the vocabulary. Basically, Skip-gram tries to maximize the softmax probability of the inner product of the center word's vector and its context word's vectors.\footnote{Mikolov et al. \cite{mikolov2013distributed} also use hierarchical softmax and negative sampling to speed up the learning process. We focus on the original simple formulation.}

\begin{figure}[t]
\centering
\includegraphics[scale=0.75]{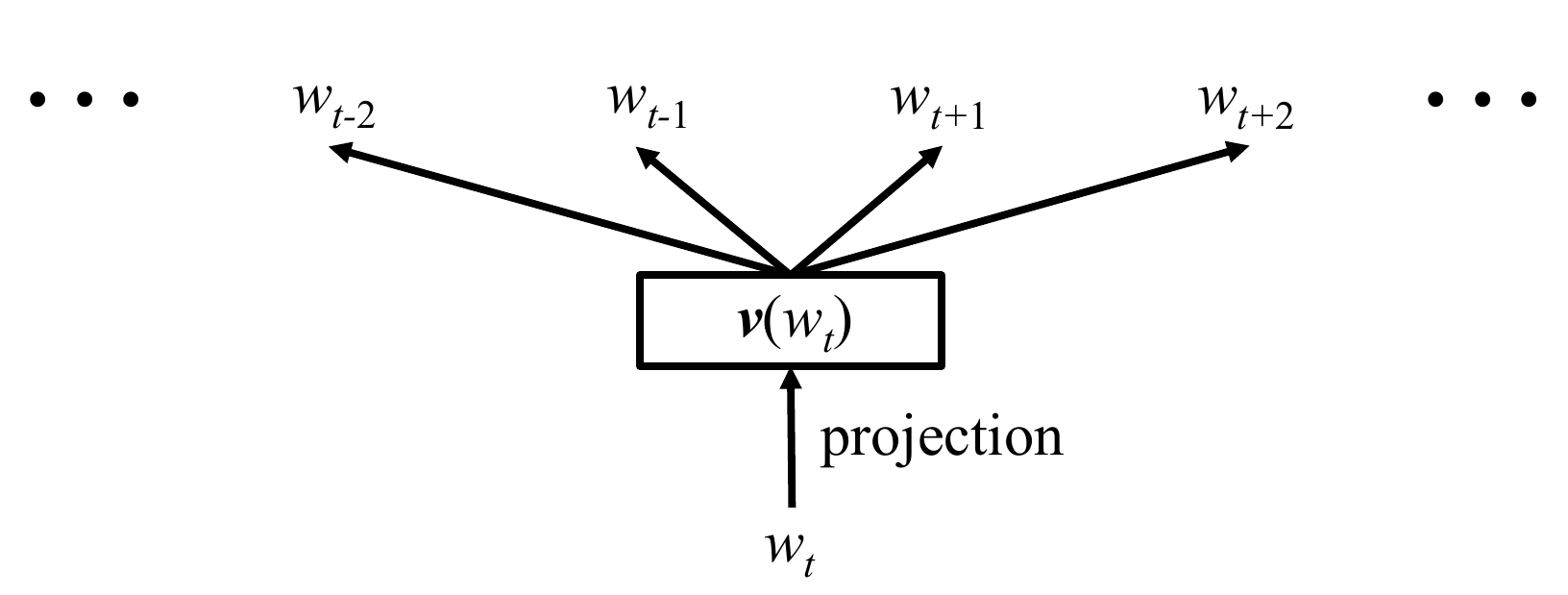}
\label{fig:skipgram}
\caption{Skip-gram model architecture: $\vb(w_t)$ is a vector representation for the word $w_t$. The goal of Skip-gram is to learn vector representations of words that are good at predicting neighboring words.}
\end{figure}

Pennington et al. proposed GloVe, \cite{pennington2014glove} which learns another word representations by using a similar principle as Skip-gram. GloVe uses the global word co-occurrence matrix to learn the word representations. Since the global co-occurrence matrix is often sparse, GloVe can be computationally less demanding than Skip-gram, which is a neural network model using the sliding context window. On the other hand, GloVe employs a weighting function that could require a considerable amount tuning effort.

Beyond one level representation like Skip-gram and GloVe, researchers also proposed hierarchical learning representations for the text corpus, which has some analogy to our healthcare setting with two level concepts namely: codes and visits. 
Le and Mikolov \cite{le2014distributed} proposes to learn representations for paragraphs and words simultaneously by treating paragraphs as one of the words. However, their algorithm assigns a fixed set of vectors for both words and paragraphs in the training data. 
Moreover, their approach does not capture the sequential order among paragraphs.
Skip-thought \cite{kiros2015skip} proposes an encoder-decoder structure: an encoder (Gated Recurrent Units (GRU) in their case) learns a representation for a sentence that is able to regenerate its surrounding sentences (via GRU again). 
Skip-thought cannot be applied directly to EHR data because unlike words in sentences, the codes in a visit are unordered. Also, the interpretation of Skip-thought model is difficult, as they rely on complex RNNs.

\subsection{Representation learning in healthcare}

Recently researchers start to explore the possibility of efficient representation learning in the medical domain. 

\paragraph{Medical text analysis}
Minarro et al. \cite{minarro2013exploring} learns the representations of medical terms by applying Skip-gram to various medical text collected from PubMed, Merck Manuals, Medscape and Wikipedia. 
De Vine et al. \cite{de2014medical} learns the representations of UMLS concepts from free-text patient records and medical journal abstracts. They first replaced the words in documents to UMLS concepts, then applied Skip-gram to learn the distributed representations of the concepts. 
However, none of them studied longitudinal EHR data with a large number of medical codes.

\paragraph{Structured visit records analysis}
Choi et al. \cite{choi2016medical}, and Choi et al. \cite{choi2016learning} both learned the distributed representation of medical codes (\textit{e.g.} diagnosis, medication, procedure codes) from structured longitudinal visit records of patients using Skip-gram. In addition, the authors demonstrated that simply aggregating the learned representation of medical codes to create a visit representation leads to improved predictive performance. However, simply aggregating the code representations is not the optimal method to generate a visit representation as it completely ignores the temporal relations across adjacent visits. We believe that taking advantage of the two-level information (the co-occurrence of codes within a visit and the sequential nature of visits) and the demographic information of patients will give us better representation for both medical codes and patient visits.

Choi et al. \cite{choi2015doctor} trained a recurrent neural networks (RNN) model to analyze the longitudinal patient records in a temporal order, and predict the diagnosis and medication codes the patient will receive in the future. 
In \cite{choi2015doctor}, the hidden layer of the RNN can be seen as the representation of the patient status over time. However, despite its outstanding performance, RNNs are difficult to interpret.

\section{Method}
\label{sec:method}
In this section, we describe the proposed algorithm \mname. We start by mathematically formulating the EHR data structure and our goal. Then we describe our approach in a top-down fashion. We also explain how to interpret the learned representations. We conclude this section with complexity analysis.

\paragraph{EHR structure and our notation}
We denote the set of all medical codes $c_1, c_2, \ldots, c_{|\mathcal{C}|}$ in our EHR dataset by $\mathcal{C}$ with size $|\mathcal{C}|$. EHR data for each patient is in the form of a sequence of visits $V_1, \ldots, V_T$ where each visit contains a subset of medical codes $V_t \subseteq \mathcal{C}$. Without loss of generality, all algorithms will be presented for a single patient to avoid cluttered notations. 
The goal of \mname is to learn two types of representations:
\begin{description}
\item[Code representations] We aim to learn an embedding function $f_C: \mathcal{C}\mapsto \mathbb{R}_+^m$ that maps every code in the set of all medical codes $\mathcal{C}$ to non-negative real-valued vectors of dimension $m$. The non-negativity constraint is introduced to improve interpretability, as discussed in details in Section \ref{ssec:interpretation}.
\item[Visit representations] Our second task is to learn another embedding function $f_V: \mathcal{V}\mapsto \mathbb{R}^n$ that maps every visit (a set of medical codes) to a real-valued vector of dimension $n$. The set $\mathcal{V}$ is the power set of the set of codes $\mathcal{C}$.
\end{description}

\subsection{Med2Vec architecture}
\label{ssec:architecture}

\begin{figure}[t]
\centering
\includegraphics[scale=0.65]{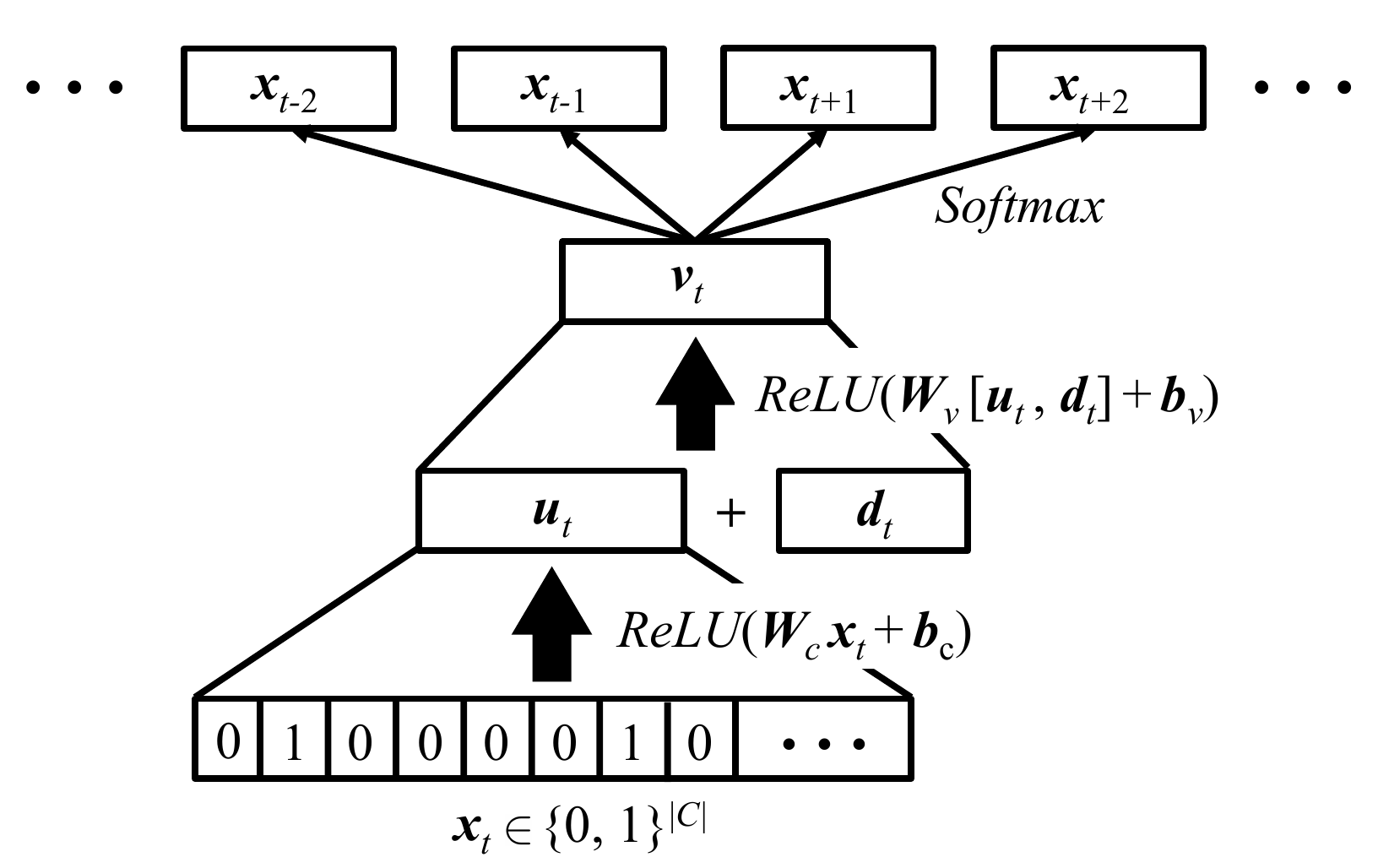} 
\caption{Structure of \texttt{Med2Vec}: A visit comprised of several medical codes is converted to a binary vector $\xb_t \in \{0,1\}^{|\mathcal{C}|}$. The binary vector is then converted to an intermediate visit representation ${\ub}_t$. ${\ub}_t$ is concatenated with a vector of demographic information $\db_t$, and converted to the final visit representation $\vb_t$, which is trained to predict its neighboring visits $\ldots, \xb_{t-2}, \xb_{t-1}, \xb_{t+1}, \xb_{t+2}, \ldots$}
\label{fig:Med2Vec}
\end{figure}

Figure \ref{fig:Med2Vec} depicts the architecture of \mname. Given a visit $V_t$, we use a multi-layer perceptron (MLP) to to generate the corresponding visit representation $\vb_t$. First, visit $V_t$ is represented by a binary vector $\xb_t \in \{0,1\}^{|\mathcal{C}|}$ where the $i$-th entry is 1 only if $c_i \in V_t$. Then $\xb_t$ is converted to an intermediate visit representation ${\ub}_t \in \mathbb{R}^{m}$ as follows,
\begin{align}
{\ub}_t = ReLU(\Wb_c \xb_{t} + \bb_c) \label{eq:intermediate} 
\end{align}
using the code weight matrix $\Wb_c \in \mathbb{R}^{m \times |\mathcal{C}|}$ and the bias vector $\bb_c \in \mathbb{R}^{m}$. The rectified linear unit is defines as $ReLU(\vb) = \max(\vb, \mathbf{0})$. Note that $\max()$ applies element-wise to vectors. We use the rectified linear unit (ReLU) as the activation function to enable interpretability, which will be discussed in section \ref{ssec:code}.

We concatenate the demographic information $\db_t \in \mathbb{R}^{ d}$, where $d$ is the size of the demographic information vector, to the intermediate visit representation ${\ub}_t$ and create the final visit representation $\vb_t \in \mathbb{R}^{n}$ as follows, 
\begin{align*}
\vb_t = ReLU(\Wb_v [\ub_t,\db_t] + \bb_v)
\end{align*}
using the visit weight matrix $\Wb_v \in \mathbb{R}^{n \times (m+d)}$ and the bias vector $\bb_v \in \mathbb{R}^{n}$, where $n$ is the predefined size of the visit representation. We use ReLU once again as the activation function. We discuss our efficient training procedure of the parameters $\Wb_c, \bb_c, \Wb_v$ and $\bb_v$ in the next subsection.

\subsection{Learning from the visit-level information}
\label{ssec:visit}
As mentioned in the introduction, the sequential information of visits can be exploited for learning efficient representations of visits and potentially codes. We train the MLP using a very straightforward intuition as follows: a visit describes a state in a continuous process that is a patient's clinical experience. Therefore, given a visit representation, we should be able to predict what has happened in the past, and what will happen in the future. Specifically, given a visit representation $\vb_t$, we train a softmax classifier that predicts the medical codes of the visits within a context window. We minimize the cross entropy error as follows,
\vspace{-2mm}
\small
\begin{align}
&\min_{\Wb_s, \bb_s} \frac{1}{T} \sum_{t=1}^{T} \sum_{-w \le i \le w, i\neq0} -{\xb_{t+i}}^{\top} \log\hat{\yb}_{t} - (\bm{1} - \xb_{t+i})^{\top}\log(\bm{1} - \hat{\yb}_{t}), \label{eq:obj_visit} \\
&\text{where}\qquad\hat{\yb}_{t} = \frac{\exp (\Wb_s \vb_{t} + \bb_s)}{\sum_{j=1}^{|\mathcal{C}|} \exp (\Wb_s[j,:] \vb_{t} + \bb_s[j])} \nonumber
\end{align}
\normalsize
where $\Wb_s \in \mathbb{R}^{|\mathcal{C}| \times n}$ and $\bb_s \in \mathbb{R}^{|\mathcal{C}|}$ are the weight matrix and bias vector for the softmax classifier, $w$ the predefined context window size, $\exp$ the element-wise exponential function, and $\bm{1}$ denotes an all one vector. We have used MATLAB's notation for selecting a row in $\Wb_s$ and a coordinate of $\bb_s$. 

\subsection{Learning from the code-level information}
\label{ssec:code}
As we described in the introduction, healthcare datasets contain two-level information: visit-level sequence information and code-level co-occurrence information. Since the loss function in Eq. \eqref{eq:obj_visit} can efficiently capture the sequence level information, now we need to find a way to use the second source of information, i.e., the intra-visit co-occurrence of the codes.


A natural choice to capture the code co-occurrence information is to use Skip-gram.  The main idea would be that the representations for the codes that occur in the same visit should predict each other. To embed Skip-gram in \mname, we can train $\Wb_c \in \mathbb{R}^{m \times |\mathcal{C}|}$ (which also produces intermediate visit level representations) so that the $i$-th column of $\Wb_c$ will be the representation for the $i$-th medical code among total $|\mathcal{C}|$ codes.  Note that given the unordered nature of the codes inside a visit,  unlike the original Skip-gram, we do not distinguish between the ``input'' medical code and the ``output'' medical code. In text, it is sensible to assume that a word can serve a different role as a center word and a context word, whereas in EHR datasets, we cannot classify codes as center or context codes. 
It is also desirable to learn the representations of different types of codes (\textit{e.g.} diagnosis, medication, procedure code) in the same latent space so that we can capture the hidden relationships between them.

However, precise interpretation of Skip-gram codes will be difficult as $\Wb_c$ will have positive and negative values. For intuitive interpretation, we should learn code representations with non-negative values. Note that in Eq.\eqref{eq:intermediate}, if the binary vector $\xb_t$ is a one-hot vector, then the intermediate visit representation ${\ub}_t$ becomes a code representation. Therefore, using the Skip-gram algorithm, we train the non-negative weight $ReLU(\Wb_c)$ instead of $\Wb_c$.
This will not only use the intra-visit co-occurrence information, but also guarantee non-negative code representations. Moreover, ReLU produces sparse code representations, which further facilitates easier interpretation of the codes.

The code representations to be learned is denoted as a matrix ${\Wb}_c' = ReLU(\Wb_c) \in \mathbb{R}^{m \times |\mathcal{C}|}$. 
From a sequence of visits $V_1, V_2, \ldots, V_T$, the code-level representations can be learned by maximizing the following log-likelihood,
\begin{align}
\min_{\Wb_c'}&\quad \frac{1}{T} \sum_{t=1}^{T} \sum_{i: c_i \in V_t} \sum_{~j: {c_j} \in V_t, j \neq i} \log p(c_j | c_i), \label{eq:obj_code} \\
\text{where}\quad p(c_j | c_i) &= \frac{\exp \Big( {\Wb}_c'[:,j]^{\top} {\Wb}_c'[:,i] \Big) }{\sum_{k=1}^{|\mathcal{C}|} \exp \Big( {\Wb}_c'[:,k]^{\top} {\Wb}_c'[:,i] \Big) }. \label{eq:relu_softmax}
\end{align}

 \vspace{-3mm}
\subsection{Unified training }
\label{ssec:unified}
The single unified framework can be obtained by adding the two objective functions \eqref{eq:obj_code} and \eqref{eq:obj_visit} as follows,
\begin{align*}
&\argmin_{\Wb, \bb} \frac{1}{T} \sum_{t=1}^{T} \Big\{ - \sum_{i: c_i \in V_t} \sum_{~j: {c_j} \in V_t, j \neq i} \log p(c_j | c_i) \\
& + \sum_{-w \le k \le w, k \neq 0} -\xb_{t+k}^{\top} \log\hat{\yb}_{t} - (\bm{1} - \xb_{t+k})^{\top}\log(\bm{1} - \hat{\yb}_{t}) \Big\}
\end{align*}

By combining the two objective functions we learn both code representations and visit representations from the same source of patient visit records, exploiting both intra-visit co-occurrence information as well as inter-visit sequential information at the same time.

\subsection{Interpretation of learned representations}
\label{ssec:interpretation}
While the original Skip-gram learns code representations that have interesting properties such as additivity, in healthcare we need stronger interpretability. We need to be able to associate clinical meaning to each dimension of both code and visit representations. 
Interpreting the learned representations is based on analyzing each coordinate in both code and visit embedding spaces.
\paragraph{Interpreting code representations}
If information is properly embedded into a lower dimensional non-negative space, each coordinate of the lower dimension can be readily interpreted. Non-negative matrix factorization (NMF) is a good example. Since we trained $ReLU(\Wb_c) \in \mathbb{R}^{m \times |\mathcal{C}|}$, a non-negative matrix, to represent the medical codes, we can employ a simple method to interpret the meaning of each coordinate of the $m$-dimensional code embedding space. We can find the top $k$ codes that have the largest values for the $i$-th coordinate of the code embedding space as follows,
\begin{align*}
\mathrm{argsort}(\Wb_c [i,:])[1:k]
\end{align*}
where $\mathrm{argsort}$ returns the indices of a vector that index its values in a descending order.
By studying the returned medical codes, we can view each coordinate as a disease group. Detailed examples are given in section \ref{ssec:result_interpretation}

\paragraph{Interpreting visit representations}
To interpret the learned visit vectors, we can use the same principle we used for interpreting the code representation. For the $i$-th coordinate of the $n$-dimensional visit embedding space, we can find the top $k$ coordinates of the code embedding space that have the strongest values as follows, 
\begin{align*}
\mathrm{argsort}(\Wb_v [i,:])[1:k]
\end{align*}
where we use the same $\mathrm{argsort}$ as before. Once we obtain a set of code coordinates, we can use the knowledge learned from interpreting the code representations to understand how each visit coordinate is associated with a group of diseases. This simple interpretation is possible because the intermediate visit representation ${\ub}_t$ is a non-negative vector, due to the $ReLU$ activation function.

In the experiments, we also tried to find the input vector $\xb_t$ that most activates the target visit coordinate \cite{erhan2009visualizing,le2013building}. However, the results were very sensitive to the initial value of $\xb_t$, and even averaging over multiple samples were producing unreliable results.

\subsection{Complexity analysis}
\label{ssec:complexity}
We first analyze the computational complexity of the code-level objective function Eq. \eqref{eq:obj_code}. Without loss of generality, we assume the visit records of all patients are concatenated into a single sequence of visits. Then the complexity for Eq. \eqref{eq:obj_code} is as follows,
\begin{align*}
\mathcal{O}(T\overline{M^2}|\mathcal{C}|m)
\end{align*}
where $T$ is the number of visits, $\overline{M^2}$ is the average of squared number of medical codes within a visit, $|\mathcal{C}|$ the number of unique medical codes, $m$ the size of the code representation. The $M^2$ factor comes from iterating over all possible pairs of codes within a visit. The complexity of the visit-level objective function Eq.\eqref{eq:obj_visit} is as follows,
\begin{align*}
\mathcal{O}(Tw(|\mathcal{C}|(m+n) + mn))
\end{align*}
where $w$ is the size of the context window, $n$ the size of the visit representation. The added terms come from generating a visit representation via MLP. Since size of code representation $m$ and size of visit representation $n$ generally have the same order of magnitude, we can replace $n$ with $m$. Furthermore, $m$ is generally smaller than $|\mathcal{C}|$ by at least two orders of magnitude. Therefore the overall complexity of \mname can be simplified as follows.
\begin{align*}
\mathcal{O}(T|\mathcal{C}|m(\overline{M^2} + w))
\end{align*}
Here we notice that $\overline{M^2}$ is generally larger than $w$. In our work, the average number of codes $\overline{M}$ per visit for two datasets are 7.88 and 3.19  according to Tables \ref{tab:data_stats}, respectively, whereas we select the window size $w$ to be at most $5$ in our experiments. 
Therefore the complexity of \mname is dominated by the code representation learning process, for which we use the Skip-gram algorithm. This means that exploiting visit-level information to learn efficient representations for both visits and codes does not incur much additional cost.

\section{Experiments}
\label{sec:exp}
In this section, we evaluate the performance of \mname in both public and proprietary datasets. First we describe the datasets. Then we describe evaluation strategies for code and visit representations, along with implementation details. Then we present the experiment results of code and visit representations with discussion. We conclude with convergence and scalability study. We make the source code of \mname publicly available at \url{https://github.com/mp2893/med2vec}.

\subsection{Dataset description}
\label{sec:data_desc}
\begin{table}[t]
\caption{Basic statistics of CHOA and CMS dataset.}
\label{tab:data_stats}
\vspace*{-2mm}
\centering
\begin{tabular}{l|c|c}
\textbf{Dataset} & \textbf{CHOA} & \textbf{CMS} \\
\hline
\hline
\# of patients & 550,339 & 831,210\\ 
\# of visits & 3,359,240 & 5,464,950\\
Avg. \# of visits per patient & 6.1 & 6.57\\
\hline
\# of unique medical codes & 28,840 & 21,033\\
- \# of unique diagnosis codes & 10,414 & 14,111\\
- \# of unique medication codes & 12,892 & N/A\\
- \# of unique procedure codes & 5,534 & 6,922\\
\hline
Avg. \# of codes per visit & 7.88 & 3.19\\
\hline
Max \# of codes per visit & 440 & 44\\
\hline
\begin{tabular}{@{}l@{}}
(95\%, 99\%) percentile\\ \# of codes per visit
\end{tabular}
& (22, 53) & (9, 13)\\
\hline
\end{tabular}
\vspace{-0.1in}
\end{table}
We evaluate performance of \mname on a dataset provided by Children's Healthcare of Atlanta (CHOA)\footnote{\url{http://www.choa.org/}}. We extract visit records from the dataset, where each visit contains several medical codes (\textit{e.g.} diagnosis, medication, procedure codes). The diagnosis codes follow ICD-9 codes, the medication codes are denoted by National Drug Codes (NDC), and the procedure codes follow Category I of Current Procedural Terminology (CPT). We exclude patients who had less that two visits to showcase \mname's ability to use sequential information of visits. The basic statistics of the dataset are summarized in Table \ref{tab:data_stats}. The data are fully de-identified and do not include any personal health information (PHI). 

We divide the dataset into two groups in a 4:1 ratio. The former is used to train \mname. The latter is held off for evaluating the visit-level representations, where we train models to predict visit-related labels. The details of the evaluation will be provided in the following subsections.

We also use CMS dataset, a \textbf{publicly available}\footnote{\url{https://www.cms.gov/Medicare/Quality-Initiatives-Patient-Assessment-Instruments/OASIS/DataSet.html}} synthetic EHR dataset. The basic information of CMS is also given in Table \ref{tab:data_stats}. Compared to CHOA dataset, the CMS dataset has more patients but fewer unique medical codes. The average number of codes per visit is also smaller than that of CHOA dataset. Since CMS dataset is synthetic, we use it only for testing the scalability of \mname and baseline models in section \ref{ssec:result_scalability}. 

\subsection{Evaluation Strategy of code representations}
\label{sec:design_code}
\paragraph{Qualitative evaluation by medical experts}
For a comprehensive qualitative evaluation, we perform a \textit{relatedness} test by selecting 100 most frequent diagnosis codes and their 5 closest diagnoses, medications and procedures in terms of cosine similarity. This allow us to know if the learned representations effectively capture the latent relationships among them. Two medical experts from CHOA check each item and assign \textit{related}, \textit{possible} and \textit{unrelated} labels.

\paragraph{Quantitative evaluation with baselines}
We use medical code groupers to quantitatively evaluate the code representations. Code groupers are used to collapse individual medical codes into clinically meaningful categories. For example, Clinical Classifications Software (CCS) groups ICD9 diagnosis codes into 283 categories such as tuberculosis, bacterial infection, and viral infection. 

We apply K-means clustering to the learned code representations and calculate the normalized mutual information (NMI) based on the group label of each code. We use the CCS as the ground truth for evaluating the code representation for diagnosis. For medication code evaluation, we use American Hospital Formulary Service (AHFS) pharmacologic-therapeutic classification, which groups NDC codes into 165 categories. For procedure code evaluation, we use the second-level grouping of CPT category I, which groups CPT codes into 115 categories.
Thus, we set the number of clusters $k$ to 283, 165, 115 respectively for the diagnosis, medication, procedure code evaluation, which matches the numbers of groups from individual groupers.

For baselines, we use popular methods that efficiently exploit co-occurrence information. Skip-gram (which is used in learning representations of medical concepts by \cite{choi2016learning,choi2016medical}) is trained using Eq. \eqref{eq:obj_code}. GloVe will be trained on the co-occurrence matrix of medical codes, for which we counted the codes co-occurring within a visit. Additionally, we also report well-known baselines such as singular value decomposition on the co-occurrence matrix.

\subsection{Evaluation strategy of visit representation}
\label{sec:design_visit}
We evaluate the quality of the visit representations  by performing two visit-level prediction tasks: predicting the future visit and predicting the present status. The former will evaluate a visit representation's potential effectiveness in predictive healthcare while the latter will evaluate the how well it captures the information in the given visit. The details of the two tasks are given below.

\noindent \textbf{Predicting future medical codes}: We predict the medical codes that will occur in the next visit using the visit representations. Specifically, given two consecutive visits $V_i$ and $V_j$, the medical codes $c \in V_j$ will be the target $\yb$, the medical codes $c \in V_i$ will be the input $\xb$, and we use softmax to predict $\yb$ given $\xb$.  The predictive performance will be measured by Top-$k$ Recall, which mimics the differential diagnosis conducted by doctors. We set $k=30$ to cover even the complex cases of CHOA dataset, as over 167,000 visits are assigned with more than 20 medical codes according to Table \ref{tab:data_stats}. We predict the grouped medical codes, obtained by  the medical groupers used in Section \ref{sec:design_code}.

\noindent \textbf{Predicting Clinical Risk Groups (CRG) level}: A patient's CRG level indicates his severity level. It ranges from 1 to 9, including 5a and 5b. The CRG levels can be divided into two groups: non-severe (CRG 1-5a) and severe (CRG 5b-9). Given a visit, we use logistic regression to predict the binary CRG class associated with the visit. We use Area Under The Curve (AUC) to measure the classification accuracy, as it is more robust to class imbalance in data.

\paragraph{Baselines}
For baselines, we use the following methods.

\noindent \textbf{Binary vector model (One-hot+)}: In order to compare with the raw input data, we use the binary vector $\xb_t$  as the visit representation. 

\noindent \textbf{Stacked autoencoder (SA)}: Stacked autoencoder is one of the most popular unsupervised representation learning algorithms \cite{vincent2010stacked}. Using the binary vector $\xb_t$ concatenated with patient demographic information as the input, we train a 3-layer stacked autoencoder (SA) \cite{bengio2007greedy} to minimize the reconstruction error.
The trained SA will then be used to generate visit representations. 

\noindent \textbf{Sum of Skip-gram vectors (Skip-gram+)}: We first learn the code-level representations with Skip-gram only (Eq. \eqref{eq:obj_code}). Then for the visit-level representation, we simply add the representations of the codes within the visit. This approach was proven very effective for heart failure prediction in \cite{choi2016medical}. We append patient demographic information at the end.

\noindent \textbf{Sum of GloVe vectors (GloVe+)}: We perform the same process as Skip-gram+, but use GloVe vectors instead of Skip-gram vectors. We use the recommended hyperparameter setting from \cite{pennington2014glove}.

\paragraph{Evaluation details}
We use the held-off dataset, which was \textit{not} used to learn the code and visit representations, to perform the two prediction tasks. The held-off dataset contains 672,110 visits assigned with CRG levels. In order to train the predictors, we divide the held-off data to training and testing folds with ration 4:1. Both softmax and logistic regression are trained for 10 epochs on the training fold. We perform 5-fold cross validation for each task to tune the regularization parameter. 
For all baseline models and \mname, we use age, sex and ethnicity as the demographic information in the input data. 

\subsection{Implementation and training details}
\label{sec:design_implementation}
For learning code and visit representations using \mname and all baselines, we use Adadelta \cite{zeiler2012adadelta} in a mini-batch fashion.
For Skip-gram, SA and \mname, we use 1,000 visits\footnote{for efficient computation, we preprocessed the EHR dataset so that the visit records of all patients are concatenated into a single sequence of visits.} per batch. For GloVe, we use 1,000 non-zero entries of the co-occurrence matrix per batch.
The optimization terminates after a fixed number of epochs. In section \ref{ssec:result_visit}, we show the relationship between training epochs and the performance. We also show the convergence behavior of \mname and the baselines in section \ref{ssec:result_scalability}.

\mname, Skip-gram, GloVe and SA are implemented with Theano 0.7.0 \cite{bergstra2010theano}. K-means clustering for the code-level evaluation and SVD are performed using Scikit-learn 0.14.1. Softmax and logistic regression models for the visit-level evaluation are implemented with Keras 0.3.1, and trained for 10 epochs. All tasks are executed on a machine equipped with Intel Xeon E5-2697v3, 256GB memory and two Nvidia K80 Tesla cards.
 
We train multiple models using various hyperparameter settings. For all models we vary the size of the code representations $m$ (or the size of the hidden layer for SA), and the number of training epochs. Additionally for \mname, we vary the size of the visit representations $n$, and the size of the visit context window $w$. 

To alleviate the curse of dimensionality when training the softmax classifier (Eq.\eqref{eq:obj_visit}) of \mname, we always use the medical code groupers of section \ref{sec:design_code} so that the softmax classifier is trained to predict the grouped medical codes instead of the exact medical codes. To confirm the impact of this strategy, we train an additional \mname without using the medical code groupers.

\subsection{Results of the code-level evaluation}
\label{ssec:result_code}
\begin{table}
\vspace*{3mm}
\caption{Average score of the medical codes from the relatedness test. 2 was assigned for \textit{related}, 1 for \textit{possible} and 0 for \textit{unrelated}}
\label{tab:qualitative_code_eval}
\vspace*{-2mm}
\centering
\begin{tabular}{c|ccc}
\hline
Average & Diagnosis & Medication & Procedure\\ 
\hline
1.34 & 1.59 & 0.95 & 1.47\\
\hline
\end{tabular}
\end{table}
\vspace{3mm}
\begin{table}
\caption{Clustering NMI of the diagnosis, medication and procedure code representations of various models. All models learned 200 dimensional code vectors. All models except SVD were trained for 10 epochs.}
\label{tab:quantitative_code_eval}
\vspace*{-2mm}
\centering
\begin{tabular}{l|c|c|c}
\hline
Model & Diagnosis & Medication & Procedure \\ 
\hline
SVD ($\sigma \Vb^{\top}$) & 0.1824 & 0.0843 & 0.1781 \\
Skip-gram & 0.2251 & 0.1216 & 0.2432 \\
GloVe & 0.4205 & 0.2163 & 0.3499 \\
Med2Vec & 0.2328 & 0.1089 & 0.21 \\
\hline
\end{tabular}
\vspace{-0.1in}
\end{table}
\begin{figure*}
\centering
\includegraphics[scale=0.5]{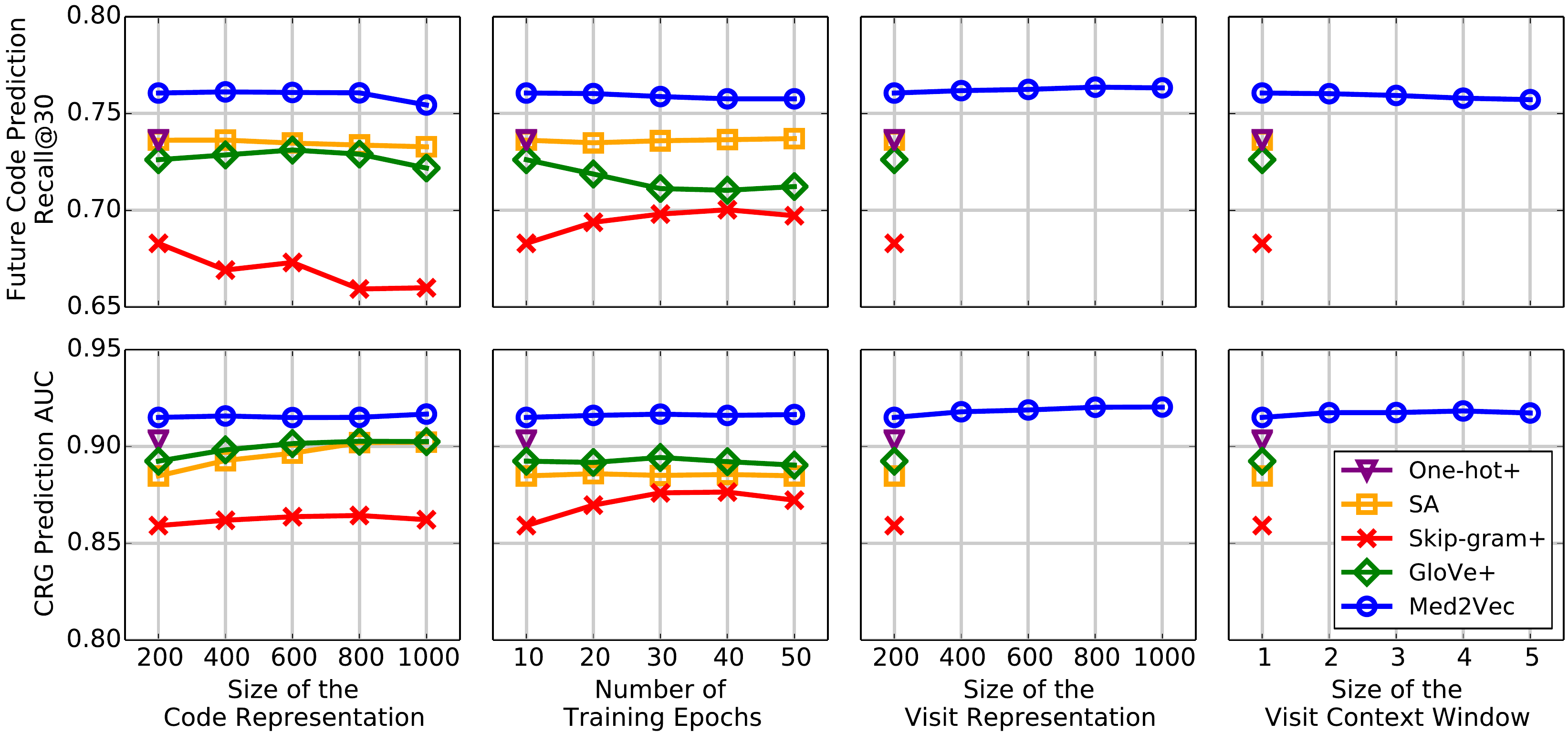}
\caption{The top row and the bottom row respectively show the Recall@30 for predicting the future medical codes and the AUC for predicting the CRG class when changing different hyperparameters. The basic configuration for Med2Vec is $m,n=200$, $w=1$, and the training epoch set to 10. The basic configuration for all baseline models is 200 for code representation size (or hidden layer size) and training epoch also set to 10. In each column, we change one hyperparameter while fixing others to the basic configuration.}
\label{fig:visit_eval}
\centering
\includegraphics[scale=0.5]{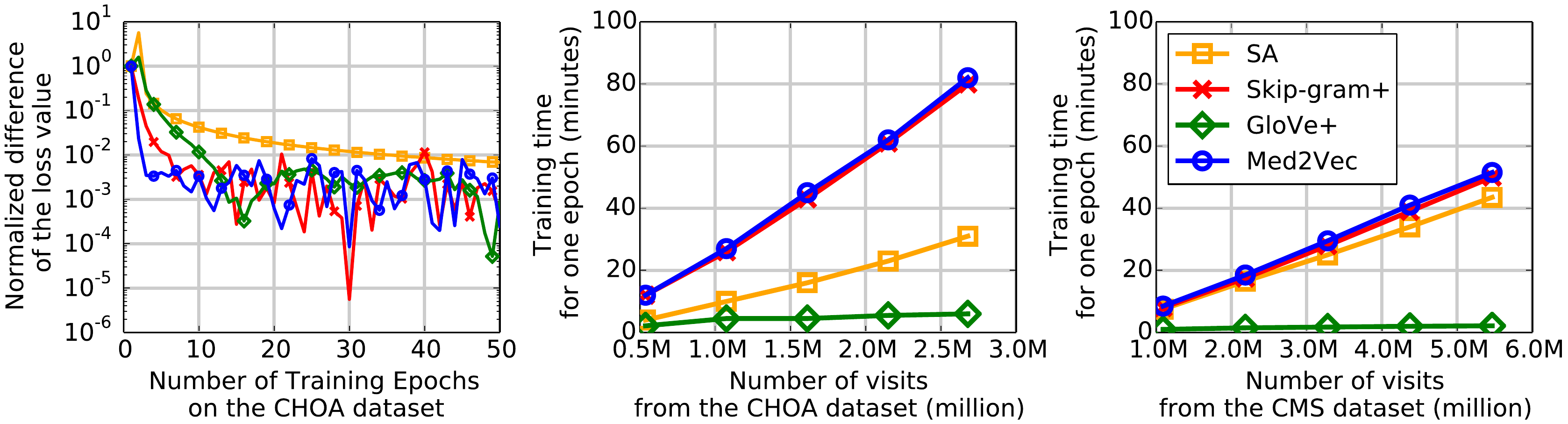}
\caption{The first figure shows the convergence behavior of all models on the CHOA dataset. The second and third figures show the relationship between the training time and the dataset size for all models respectively using the CHOA dataset and the CMS dataset.}
\label{fig:time_eval}
\end{figure*}
Table \ref{tab:qualitative_code_eval} shows the average score of the medical codes from the qualitative code evaluation. On average, \mname successfully captures the relationship between medical codes. However, \mname seems to have a hard time capturing proper representation of medications. This is due to the precise nature the medication prescription. For example, \mname calculated that \textit{Ofloxacin}, an antibiotic sometimes used to treat middle-ear infection, was related to \textit{sensorineural hearling loss} (SNHL), an inner-ear problem. On the surface level, this is a wrong relationship. But \mname can be seen as capturing the deeper relationship between medical concepts that is not always clear on the surface level.

Table \ref{tab:quantitative_code_eval} shows the clustering NMI of diagnosis, medication and procedure codes, measured for various models. 
\mname shows more or less similar conformity to the existing groupers as Skip-gram. SVD shows the weakest conformity among all models. GloVe exhibits significantly stronger conformity than any other models. Exploiting the global co-occurrence matrix seems to help learn code representations where similar codes are closer to each other in terms of Euclidean distance.

However, the degree of conformity of the code representations to the groupers does not necessarily indicate how well the code representations capture the hidden relationships. For example, CCS categorizes ICD9 224.4 \textit{Benign neoplasm of cornea} as CCS 47 \textit{Other and unspecified benign neoplasm}, and ICD9 370.00 \textit{Unspecified corneal ulcer} as CCS 91 \textit{Other eye disorders}. But the two diagnosis codes are both eye related problems, and they could be considered related in that sense. Therefore we recommend the readers use the evaluation results for comparing the performance between \mname and other baselines, rather than for measuring the absolute performance. 

In the following visit-level evaluation, we show that the code representations' strong conformity to the existing groupers alone does not directly transfer to good visit representations.

\subsection{Results of the visit-level evaluation}
\label{ssec:result_visit}
\begin{table}
\caption{Performance comparison of two Med2Vec models. The top row was trained with the grouped code as mentioned in section \ref{sec:design_implementation}. The bottom row was trained without using the groupers. Both models were trained for 10 epochs with $m,n=200$, $w=1$.}
\label{tab:grouped_vs_full}
\centering
\begin{tabular}{l|c|c}
\hline
Model & Future code prediction & CRG prediction\\ 
\hline
Grouped codes & 0.7605 & 0.9150\\
Exact codes & 0.7574 & 0.9155\\
\hline
\end{tabular}
\end{table}
The first row of Figure \ref{fig:visit_eval} shows the Recall@30 for predicting the future medical codes. First, in all of the experiments, \mname achieves the highest performance, despite the fact that it is constrained to be positive and interpretable. The second observation is that \mname's performance is robust to choice of the hyperparameters in a wide range of values. Comparing to a more volatile performance of Skip-gram, we can see that including the visit information in training not only improves the performance, but also stabilizes it too.

Another fascinating aspect of the results is the overfitting pattern in different algorithms. Increasing the code representation size degrades the performance of all of the algorithms, as it leads to overfitting. Similar behavior can be seen as we train GloVe+ for more epochs which suggests early stopping technique should be used in representation learning \cite{bengio2009learning}.
For \mname, increasing the visit representation size $n$ seems to have the strongest influence to its predictive performance.

The bottom row of figures in Figure \ref{fig:visit_eval} shows the AUC for predicting the CRG class of the given visit. The overfitting patterns are not as prominent as the previous task. This is due to the different nature of the two prediction tasks. While the goal of CRG prediction is to predict a value related to the current visit, predicting the future codes is taking a step away from the current visit.  
This different nature of the two tasks also contributes to the better performance of One-hot+ on the CRG prediction. One-hot+ contains the entire information of the given visit, although in a very high-dimensional space. Therefore predicting the CRG level, which has a tight relationship with the medical codes within a visit, is an easier task for One-hot+ than predicting the future codes.

Table \ref{tab:grouped_vs_full} shows the performance comparison between two different \mname models. The top model is trained with the grouped codes as explained in section \ref{sec:design_implementation}, while the bottom models is trained with the exact codes. Considering the marginal difference of the CRG prediction AUC, it is evident that our strategy to alleviate the curse of dimensionality was beneficial. Moreover, using the grouped codes will improve the training speed as the softmax function will require less computation.

\subsection{Convergence behavior and scalability}
\label{ssec:result_scalability}
We first compare the convergence behavior of \mname with Skip-gram (Eq. \eqref{eq:obj_code}), GloVe and SA. For SA, we measure the convergence behavior of a single-layer. We train the models for 50 epochs and plot the normalized difference of the loss value $\frac{\mathcal{L}_t - \mathcal{L}_{t-1}}{\mathcal{L}_t}$, where $\mathcal{L}_t$ denotes the loss value at time $t$.
To study the scalability of the models, we use both CHOA dataset and CMS dataset. We vary the size of the training data and plot the time taken for each model to run one epoch.

The left figure of Fig \ref{fig:time_eval} shows the convergence behavior of all models when trained on the CHOA dataset. SA shows the most stable convergence behavior, which is natural given that we used a single-layer SA, a much less complex model compared to GloVe, Skip-gram and \mname. All models except SA seem to reach convergence after 10 epochs of training. Note that \mname shows similar, if not better convergence behavior compared to Skip-gram even with added complexity.

The center figure of Fig \ref{fig:time_eval} shows the minutes taken to train all models for one epoch using the CHOA dataset. As we have analzyed in section {ssec:complexity}, \mname takes essentially the same time to train for one epoch. Both Skip-gram and \mname, however, takes longer than SA and GloVe. This is mainly due to having the softmax function for training the code representations. GloVe, which is trained on the very sparse co-occurrence matrix naturally takes the least time to train. 

The right figure of Fig \ref{fig:time_eval} shows the training time when using the CMS dataset. Note that \mname and Skip-gram takes similar time to train as SA. This is due to the smaller number of codes per visit, which is the computationally dominating factor of both \mname and Skip-gram. GloVe takes less time as the number of unique codes are smaller in the CMS dataset. SA, on the other hand, takes more time because the number of visits have doubled while the the number of unique codes is about 73\% of that of the CHOA dataset.

\section{Interpretation}
\label{sec:interpret}
Given the importance of interpretability in healthcare, we demonstrate three stages of  interpretability for our model in collaboration with the medical experts from CHOA. First, to analyze the learned code representations we show top five medical codes for each of six coordinates of the code embedding space and explain the characteristic of each coordinate. This way, we show how we can annotate each dimension of the code embedding space with clinical concepts. The six coordinates are specifically chosen so that they can be used in the later stages. 
Second, we demonstrate the interpretability of \mname's visit representations by analyzing the meaning of two coordinates in the visit embedding space.

Finally, we extend the interpretability of \mname to a real-world task, the CRG prediction, and analyze the medical codes that have strong influence on the CRG level.
Once we learn the logistic regression weight $\wb_{LR}$ for the CRG prediction, we can extract knowledge from the learned weights by analyzing the visit coordinates to which the weights are strongly connected. 

Instead of analyzing the visit coordinates, however, we propose an approximate way of directly finding out which code coordinate plays an important role in predicting the CRG class. Our goal is to find $\ub_t$ such that maximizes the output activation as follows\footnote{As we are interested in influential codes, we assume the demographic information vector is zero vector and omit it for ease of notation.}
\begin{equation}
\ub_t^{\star} = \argmax_{\ub_t,\|\ub_t\|_2=1,\ub_t \succeq  0 }\left[ReLU(\Wb_v \ub_t+\bb_v)\right]^{\top}\wb_{LR}
\label{eq:interpretation}
\end{equation}
Given the fact that $ReLU(\cdot)$ is an increasing function (not-strictly though), we make an approximation and find the solution without the $ReLU(\cdot)$ term. The approximate solution can be found in closed form $\ub_t^{\star} \propto (\Wb_v^{\top} \wb_{LR})_+$. Finally, we calculate the element-wise product of $\ub_t^{\star}$ and $\max(\Wb_c + \bb_c)$. This is to take into account the fact that each code coordinate has different maximum value. Therefore, instead of simply selecting the code coordinate with the strongest connection to the CRG level, we consider each coordinate's maximum ability to activate the positive CRG prediction. 

The resulting vector will show the maximum influence each code coordinate can have on the CRG prediction. 

\subsection{Results}
\label{ssec:result_interpretation}
\begin{table*}[t]
\caption{Medical codes with the strongest value in six different coordinates of the 200 dimensional code embedding space. We choose ten medical codes per coordinate. Shortened descriptions of diagnosis codes are compensated by their ICD9 codes. Medications and procedures are appended with (R) and (P) respectively.}
\label{tab:code_interpretation}
\scriptsize
\centering
\begin{tabular}{|l|l|l|}
\hline
\multicolumn{1}{|c|}{Coordinate 112} & \multicolumn{1}{c|}{Coordinate 152} & \multicolumn{1}{c|}{Coordinate 141} \\
\hline
\begin{tabular}{@{}l@{}}
Kidney replaced by transplant (V42.0)\\
Hb-SS disease without crisis (282.61)\\
Heart replaced by transplant (V42.1)\\
RBC antibody screening (P)\\
Complications of transplanted \\bone marrow (996.85)\\
Sickle-cell disease (282.60)\\
Liver replaced by transplant (V42.7)\\
Hb-SS disease with crisis (282.62)\\
Prograf PO (R)\\
Complications of transplanted heart \\(996.83)\\
\end{tabular}
&
\begin{tabular}{@{}l@{}}
X-ray, knee (P)\\
X-ray, thoracolumbar (P)\\
Accidents in public building (E849.6)\\
Activities involving gymnastics (E005.2)\\
Struck by objects/persons in sports (E917.0)\\
Encounter for removal of sutures (V58.32)\\
Struck by object in sports (E917.5)\\
Unspecified fracture of ankle (824.8)\\
Accidents occurring in place for \\recreation and sport (E849.4)\\
Activities involving basketball (E007.6)\\
\end{tabular}
&
\begin{tabular}{@{}l@{}}
Cystic fibrosis (277.02)\\
Intracranial injury (854.00)\\
Persistent mental disorders (294.9)\\
Subdural hemorrhage (432.1)\\
Neurofibromatosis (237.71)\\
Other conditions of brain (348.89)\\
Conductive hearing loss (389.05)\\
Unspecified causes of encephalitis, \\myelitis, encephalomyelitis (323.9)\\
Sensorineural hearing loss (389.15)\\
Intracerebral hemorrhage (431)\\
\end{tabular}
\\
\hline
\multicolumn{1}{|c|}{Coordinate 184} & \multicolumn{1}{c|}{Coordinate 190} & \multicolumn{1}{c|}{Coordinate 199} \\
\hline
\begin{tabular}{@{}l@{}}
Pain in joint, shoulder region (719.41)\\
Pain in joint, lower leg (719.46)\\
Pain in joint, ankle and foot (719.47)\\
Pain in joint, multiple sites (719.49)\\
Generalized convulsive epilepsy (345.10)\\
Pain in joint, upper arm (719.42)\\
Cerebral artery occlusion (434.91)\\
MRI, brain (780.59)\\
Other joint derangement (718.81)\\
Fecal occult blood (790.6)\\
\end{tabular}
&
\begin{tabular}{@{}l@{}}
Down's syndrome (758.0)\\
Congenital anomalies (759.89)\\
Tuberous sclerosis (759.5)\\
Anomalies of larynx, trachea, \\and bronchus (748.3)\\
Autosomal deletions (758.39)\\
Conditions due to anomaly of \\unspecified chromosome (758.9)\\
Acquired hypothyroidism (244.9)\\
Conditions due to chromosome \\anomalies (758.89)\\
Anomalies of spleen (759.0)\\
Conditions due to autosomal \\anomalies (758.5)\\
\end{tabular}
&
\begin{tabular}{@{}l@{}}
Infantile cerebral palsy (343.9)\\
Congenital quadriplegia (343.2)\\
Congenital diplegia (343.0)\\
Quadriplegia (344.00)\\
Congenital hemiplegia (343.1)\\
Baclofen 10mg tablet (R)\\
Wheelchair management (P)\\
Tracheostomy status (V44.0)\\
Paraplegia (344.1)\\
Baclofen 5mg/ml liquid (R)\\
\end{tabular}
\\
\hline
\end{tabular}
\end{table*}
Table \ref{tab:code_interpretation} shows top ten codes with the largest value in each of the six coordinates of the code embedding space. The coordinate 112 is clearly related to sickle-cell disease and organ transplant. The two are closely related in that sickle cell disease can be treated with bone-marrow transplant. Prograf is a medication used for preventing organ rejection.
Coordinate 152 groups medical codes related to sports-related injuries, specifically broken bones. Coordinate 141 is related to brain injuries and hearing loss due to the brain injuries. Neurofibromatosis(NF) is also related to this coordinate because it can cause tumors along the nerves in the brain. Cystic fibrosis(CF) seems to be a weak link in this group as it is only related to NF in the sense that both NF and CF are genetically inherited. 
Coordinate 184 clearly represents medical codes related to epilepsy. Epilepsy is often accompanied by convulsions, which can cause joint pain. Cerebral artery occlusion is related epilepsy in the sense that epileptic seizures can be a manifestation of cerebral arterial occlusive diseases\cite{cocito1982epileptic}. Also, both blood in feces and the joint pain can be attributed to Henoch--Sch\"onlein purpura, a disease primarily found in children.
Coordinate 190 groups diseases that are caused by congenital chromosome anomalies, especially the autosome. Acquired hypothyroidism seems to be an outlier of this coordinate. 
Coordinate 199 is strongly related to congenital paralysis. Baclofen is a medication used as a muscle relaxer. Quadraplegia patients can have weakened respiratory function due to impaired abdominal muscles\cite{forner1980lung}, in which case tracheostomy could be required.

We now analyze two visit coordinates: coordinate 50 and 41. Both visit coordinates have the strongest connection to the logistic regression learned for the  CRG prediction. For visit coordinate 50, the two strongest code coordinates connected to it are code coordinates 112 and 152. Then naturally, from our analysis above, we can easily see that visit coordinate 50 is strongly activated by sickle-cell disease and sports-related injuries. For visit coordinate 41, code coordinates 141 and 184 have the strongest connection. Again from the analysis above, we can directly infer that visit coordinate 41 can be seen as a patient group consisting of brain damage \& hearing loss patients and epilepsy patients. By repeating this process, we can find the code coordinates that are likely to strongly influence the CRG level.

However, finding the influential code coordinates for CRG level can be achieved without analyzing the visit representation if we use Eq.\eqref{eq:interpretation}. Applying Eq.\eqref{eq:interpretation} to the logistic regression weight of the CRG prediction, we learned that code coordinates 190 and 199 are the two strongest influencer of the CRG level. Using the analysis from above, we can naturally conclude that patients suffering from congenital chromosome anomalies or congenital paralysis are most likely to be considered to be in severe states, which is obviously true in any clinical setting. 

\section{Conclusion}
\label{sec:conclusion}
In this paper, we proposed \mname, a scalable two layer neural network for learning lower dimensional representations for medical concepts. \mname incorporates both code co-occurence information and visit sequence information of the EHR data which improves the accuracy of both code and visit representations.  Throughout several experiments, we successfully demonstrated the superior performance of \mname in two predictive tasks and provided clinical interpretation of the learned representations.
 

\section{Acknowledgments}
This work was supported by the National Science Foundation, award IIS- \#1418511 and CCF-\#1533768, Children's Healthcare of Atlanta, CDC I-SMILE project, Google Faculty Award,  AWS Research Award, Microsoft Azure Research Award and UCB.

%
\bibliographystyle{abbrv}
\bibliography{med2vec}  
%
%
\end{document}